\pdfoutput=1

\documentclass[11pt]{article}

\usepackage[preprint]{acl}
\usepackage{dsfont} 
\usepackage{times}
\usepackage{latexsym}
\usepackage{bbm} 
\usepackage{bm} 
\usepackage{enumitem}
\usepackage{booktabs}
\usepackage{comment}
\usepackage[T1]{fontenc}

\usepackage[utf8]{inputenc}

\usepackage[font=small,labelfont=bf]{caption}

\usepackage{microtype}

\usepackage{inconsolata}
\usepackage{amsmath}
\usepackage{amssymb}
\usepackage{amsfonts}
\usepackage{amsthm}

\usepackage{thmtools}
\usepackage{thm-restate}

\usepackage{graphicx}
\usepackage{todonotes}
\definecolor{mintgreen}{RGB}{152, 255, 152}
\makeatletter
\newcommand*\iftodonotes{\if@todonotes@disabled\expandafter\@secondoftwo\else\expandafter\@firstoftwo\fi}  %
\makeatother

\newcommand{\mymacro}[1]{#1}

\newcommand{\defn}[1]{\textbf{#1}}
\newcommand{\defeq}{\mathrel{\stackrel{\textnormal{\tiny def}}{=}}}

\newcommand{\R}{\mathbb{R}}

\newcommand{\tempact}{\mymacro{h}}
\newcommand{\paststeps}{\mymacro{l}}
\usepackage{xcolor}
\newif\ifshowrevision          
\showrevisiontrue              

\ifshowrevision
  
\else
  
\fi

\newcommand{\predictors}{\mymacro{\mathbf{x}_{m}}}
\newcommand{\tpredictors}{\mymacro{\mathbf{x}_{m}^{\top}}}

\newcommand{\weights}{\mymacro{\mathbf{w}}}
\newcommand{\tempkern}{\mymacro{\phi}}

\newcommand{\dinf}{\mathrm{d}}

\newcommand{\sequence}{{\mymacro{\ensuremath{\mathcal{T}}}}}

\newcommand{\location}{{\mymacro{\textbf{s}}}}
\newcommand{\nloc}{{\mymacro{\location_n}}}

\newcommand{\duration}{{\mymacro{d}}}

\newcommand{\timestamp}{\mymacro{t}}
\newcommand{\ntime}{\mymacro{\timestamp_n}}
\newcommand{\nptime}{\mymacro{\timestamp_{n-1}}}
\newcommand{\ndur}{\mymacro{\duration_{n}}}

\newcommand{\nhist}{\mymacro{\history_{n}}}
\newcommand{\nphist}{\mymacro{\history_{n-1}}}

\newcommand{\intensity}{\mymacro{\lambda}}
\newcommand{\baseintensity}{\mymacro{\nu}}

\newcommand{\history}{\mymacro{\ensuremath{\mathcal{H}}}}
\newcommand{\screen}{\mymacro{\ensuremath{\Omega}}}

\newcommand{\alphahat}{\mymacro{\bm{\widehat{\alpha}}}}
\newcommand{\alphaest}{\mymacro{{h(\mathbf{x}_m^\top \alphahat )}}}
\newcommand{\betahat}{\mymacro{{\bm{\widehat{\beta}}}}}
\newcommand{\betaest}{\mymacro{{h(\mathbf{x}_m^\top \betahat)}}}

\newcommand{\spilledpred}{\mymacro{x_{mk}}}
\newcommand{\npdur}{\mymacro{\duration_{n-1}}}
\newcommand{\mdur}{\mymacro{\duration_{m}}}

\usepackage{mathtools}
\usepackage{ellipsis}
\makeatletter
\renewcommand*{\mathellipsis}{%
  \mathinner{%
    \kern\ellipsisbeforegap%
    {\ldotp}\kern\ellipsisgap%
    {\ldotp}\kern\ellipsisgap%
    {\ldotp}\kern\ellipsisaftergap%
  }%
}
\renewcommand*{\dotsb@}{%
  \mathinner{%
    \kern\ellipsisbeforegap%
    {\cdotp}\kern\ellipsisgap%
    {\cdotp}\kern\ellipsisgap%
    {\cdotp}\kern\ellipsisaftergap%
  }%
}
\renewcommand*{\@cdots}{%
  \mathinner{%
    \kern\ellipsisbeforegap%
    {\cdotp}\kern\ellipsisgap%
    {\cdotp}\kern\ellipsisgap%
    {\cdotp}\kern\ellipsisaftergap%
  }%
}
\renewcommand*{\ellipsis@default}{%
  \ellipsis@before
  \kern\ellipsisbeforegap
  .\kern\ellipsisgap
  .\kern\ellipsisgap
  .\kern\ellipsisgap
  \ellipsis@after\relax}
\renewcommand*{\ellipsis@centered}{%
  \ellipsis@before
  \kern\ellipsisbeforegap
  .\kern\ellipsisgap
  .\kern\ellipsisgap
  .\kern\ellipsisaftergap
  \ellipsis@after\relax}
\AtBeginDocument{%
  \DeclareRobustCommand*{\dots}{%
    \ifmmode\@xp\mdots@\else\@xp\textellipsis\fi}}
\def\ellipsisgap{-.05em}
\def\ellipsisbeforegap{-.05em}
\def\ellipsisaftergap{.05em}
\makeatother

\usepackage[normalem]{ulem} 

\definecolor{ETHBlue}{RGB}{33,92,175}   
\definecolor{ETHGreen}{RGB}{98,115,19}      
\definecolor{ETHPurpleDark}{RGB}{140,10,89} 
\definecolor{ETHPurple}{RGB}{163,7,116} 
\definecolor{ETHGray}{RGB}{111,111,111} 
\definecolor{ETHRed}{RGB}{183,53,45}    
\definecolor{ETHPetrol}{RGB}{0,120,148} 
\definecolor{ETHBronze}{RGB}{142,103,19}    
\definecolor{ETHOrange}{RGB}{230, 100, 50}

\colorlet{MacroColor}{ETHGreen}
\colorlet{MACROCOLOR}{MacroColor}

\definecolor{WildStrawberry}{rgb}{1.0, 0.26, 0.64}
\definecolor{Wisteria}{rgb}{0.79, 0.63, 0.86}
\definecolor{Olive}{rgb}{0.5, 0.5, 0.0}
\definecolor{Plum}{rgb}{0.56, 0.27, 0.52}
\definecolor{NavyBlue}{rgb}{0.0, 0.0, 0.5}
\definecolor{OliveGreen}{rgb}{0.33, 0.42, 0.18}
\definecolor{RoyalPurple}{rgb}{0.47, 0.32, 0.66}

\usepackage[capitalize,noabbrev]{cleveref}
\crefname{section}{\S}{\S\S}
\crefname{table}{Tab.}{Tab.}
\crefname{figure}{Fig.}{Figs.}
\crefname{algorithm}{Alg.}{}
\crefname{equation}{Eq.}{Eqs.}
\crefname{appendix}{App.}{Apps.}
\crefname{theorem}{Theorem}{}
\crefname{prop}{Proposition}{}
\crefname{definition}{Def.}{}
\crefname{cor}{Corollary}{}
\crefname{observation}{Observation}{}
\crefname{assumption}{Assumption}{}
\crefname{hyp}{Hyp.}{Hypotheses}
\crefformat{section}{\S#2#1#3}
\crefname{namedtheorem}{Hyp.}{Hypotheses}

\crefformat{equation}{\eqA Eq.~\eqB #2#1#3)}
\newcommand{\eqA}{}
\newcommand{\eqB}{(}
\DeclareRobustCommand{\pcref}[1]{%
  \begingroup
  \renewcommand{\eqA}{(}\renewcommand{\eqB}{}%
  \cref{#1}%
  \endgroup
}

\newcommand{\saveforcameraready}[1]{}
\newcommand{\cutforspace}[1]{}

\usepackage{stmaryrd} 

\newcommand{\eos}{{\mymacro{\textsc{eos}}}}

\DeclareTextCommandDefault{\textvisiblespace}{%
  \mbox{\kern.06em\vrule \@height.3ex}%
  \vbox{\hrule \@width.3em}%
  \hbox{\vrule \@height.3ex}}

\newcommand{\bigint}{\mymacro{\Lambda}}
\newcommand{\gammadis}{\mymacro{\gamma}}

\newcommand{\spatmean}{\mymacro{\mu_m}}
\newcommand{\spatmeanbase}{\mymacro{\mu^{\mathrm{b}}_m}}
\newcommand{\spatmeanaff}{\mymacro{\mu^{\mathrm{a}}_m}}

\newcommand{\spatmeanfull}{\mymacro{\mu^{\mathrm{f}}_m}}
\newcommand{\durmean}{\mymacro{\xi_n}}
\newcommand{\durmeanconv}{\mymacro{\xi^{\mathrm{c}}_n}}
\newcommand{\durmeanspill}{\mymacro{\xi^{\mathrm{M}}_n}}

\newcommand{\spatialparamslocation}{\mymacro{\mathbf{A}}}
\newcommand{\spatialparamsbias}{\mymacro{\mathbf{b}}}
\newcommand{\spatialparamspreds}{\mymacro{\mathbf{C}}}
\newcommand{\spilledcoeffs}{\mymacro{w_{mk}^\prime}}
\newcommand{\agamma}{\mymacro{\alpha_k}}
\newcommand{\bgamma}{\mymacro{\beta_k}}
\newcommand{\thgamma}{\mymacro{\theta_k}}

\newcommand{\durpredictors}{\mymacro{\mathbf{x}_n}}
\newcommand{\tdurpredictors}{\mymacro{\mathbf{x}_n^{\top}}}

\newcommand{\alphapar}{\mymacro{\bm{\alpha}}}
\newcommand{\betapar}{\mymacro{\bm{\beta}}}

\newcommand{\adjterm}{\mymacro{\delta}}
\newcommand{\spatdens}{\mymacro{\psi}}
\newcommand{\mtime}{\mymacro{\timestamp_{m}}}
\newcommand{\mloc}{\mymacro{\location_{m}}}

\newcommand{\unit}{{\mymacro{w}}}

\newcommand{\alphabet}{{\mymacro{\Sigma}}}
\newcommand{\eosalphabet}{{\mymacro{\overline{\alphabet}}}}
\newcommand{\ctx}{{\mymacro{\boldsymbol{\unit}_{<t}}}}

\newcommand{\hawksdensity }{\mymacro{f}}
\newcommand{\durdensity}{{\mymacro{g}}}

\title{A Spatio-Temporal Point Process for
Fine-Grained \\ Modeling of Reading Behavior
}

\newcommand{\ethz}{1}
\newcommand{\mpi}{2}

\author{\textbf{Francesco Ignazio Re}$^{\ethz}$~\;~\;~\textbf{Andreas Opedal}$^{\ethz,\mpi}$~\;~\;~\textbf{Glib Manaiev}$^{\ethz}$\\
\textbf{Mario Giulianelli}$^{\ethz}$~\;~\;~\textbf{Ryan Cotterell}$^{\ethz}$ \\
$^{\ethz}$ETH Z{\"u}rich
    \quad $^{\mpi}$Max Planck Institute for Intelligent Systems, T{\"u}bingen \\
    \texttt{\{\href{francesco.re@inf.ethz.ch}{francesco.re}, \href{andreas.opedal@inf.ethz.ch}{andreas.opedal}, \href{ryan.cotterell@inf.ethz.ch}{ryan.cotterell}\}@inf.ethz.ch}
  }

\begin{document}
\maketitle
\begin{abstract}
Reading is a process that unfolds across space and time, alternating between fixations where a reader focuses on a specific point in space, and saccades where a reader rapidly shifts their focus to a new point.
An ansatz of psycholinguistics is that modeling a reader's fixations and saccades yields insight into their online sentence processing. 
However, standard approaches to such modeling rely on aggregated eye-tracking measurements and models that impose strong assumptions, ignoring much of the spatio-temporal dynamics that occur during reading.
In this paper, we propose a more general probabilistic model of reading behavior, based on a marked spatio-temporal point process, that captures not only \emph{how long} fixations last, but also \emph{where} they land in space and \emph{when} they take place in time. The saccades are modeled using a Hawkes process, which captures how each fixation excites the probability of a new fixation occurring near it in time and space. The duration time of fixation events is modeled as a function of fixation-specific predictors convolved across time, thus capturing spillover effects.
Empirically, our Hawkes process model exhibits a better fit to human saccades than baselines. 
With respect to fixation durations, we observe that incorporating contextual surprisal as a predictor results in only a marginal improvement in the model's predictive accuracy.
This finding suggests that surprisal theory struggles to explain fine-grained eye movements.

\vspace{.2em}
\hspace{1.25em}\includegraphics[width=1.4em,height=1.4em]{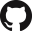}{\hspace{.75em}\parbox{\dimexpr\linewidth-2\fboxsep-2\fboxrule}{\url{https://github.com/rycolab/spatio-temporal-reading}}}

\end{abstract}

\section{Introduction}
Reading is a cognitively complex skill. As we read, our eyes move through an interdigitated sequence of \defn{fixations}, brief pauses that allow for the perception and processing of linguistic material, and \defn{saccades}, rapid movements that shift focus to the next point of interest.
A longstanding premise in psycholinguistics is that eye movements during reading provide a direct window into the cognitive processes underlying language comprehension \citep{McConkie1979,just-carpenter-1980,Rayner1989online,Findlay_Walker_1999}. 
Based on this premise, eye-tracking experiments have emerged as one of the most effective paradigms for testing and refining theories of language processing \citep{rayner1998eye,frank-etal-2013-reading}.

\begin{figure*}
    \centering
    \vspace{-8pt}
    \includegraphics[width=0.99\textwidth]{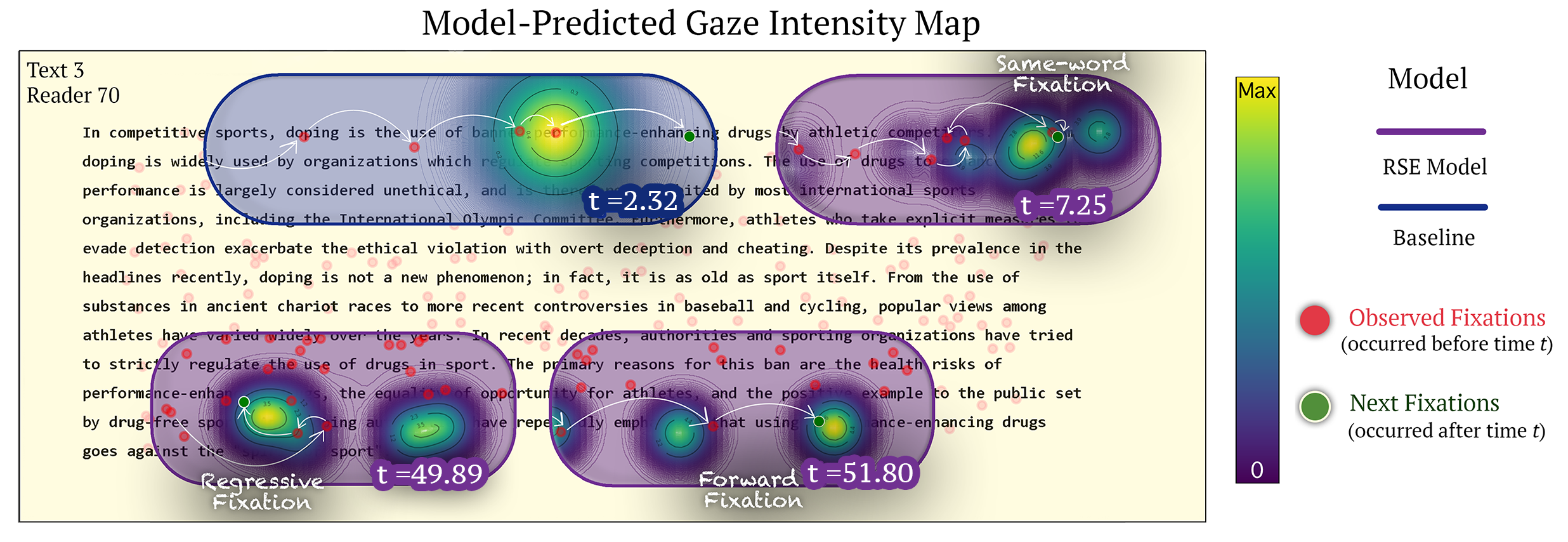}
    \vspace{-5pt}
    \caption{Illustration of how the intensity function of the saccade model evolves over time within a scanpath, and how model predictions compare to held-out data. The intensity function is shown at selected timestamps and visualizes the predicted fixation density of two models: our reader-specific effects (RSE) Hawkes process (in {\color{ETHPurpleDark!70!black} purple}), which accounts for reader-specific effects and directional saccade tendencies (e.g., forward shifts in reading), and a last-fixation baseline model (in {\color{ETHBlue!80!black} blue}), which concentrates density around the most recent fixation location; see details in \cref{sec:hawkes-process} and \cref{sec:saccade-experiments}. Red dots indicate observed fixations prior to time $\timestamp$, while green dots mark the fixation immediately following $\timestamp$. Note how the RSE model captures key reading behaviors, including forward saccades ($\timestamp = 51.80$), backward regressions ($\timestamp = 49.89$), and re-fixations on the same word ($\timestamp = 7.25$).\looseness=-1 
    }
    \label{fig:fixation_intensity}
    \vspace{-12pt}
\end{figure*}

Data collected in eye-tracking studies, termed scanpaths, consist of sequences of the participants' fixations on a text displayed on a two-dimensional coordinate space, e.g., a screen placed in front of the participant, along with the fixations' durations and onset time.
In modern computational psycholinguistic studies, such raw data is typically aggregated into summary measurements, such as total fixation duration, the summed duration of all fixations on a chosen linguistic unit, and gaze duration, the summed duration of all fixations between landing on a word and moving to another.
These summary measurements are then treated as dependent variables in a (generalized) linear model during analysis \citep{smith2013-log-reading-time,goodkind-bicknell-2018-predictive,wilcox2020}.

Such aggregation, however, is an inherently lossy process.
From a temporal perspective, combining multiple fixations in a single measurement may conflate several factors that underlie the aggregated behaviors. 
For example, total fixation time includes first fixations as well as fixations where the reader moves back to a previously fixated point (regressive saccades), which are posited to correspond to different cognitive processes \citep{wilcox-2024-regressions}.
From a spatial perspective, aggregations inherently rely on pre-defined regions of interest \citep{giulianelli-etal-2024-proper}.
Aggregating fixations, which is most commonly done at the word level,
discards any information about \emph{where} saccades land within the boundaries of a word
and hinders investigations into smaller linguistic units, e.g., syllables or morphemes. 
In sum, while aggregations help simplify the challenge of modeling and interpreting the complex spatio-temporal dynamics of fixations and saccades, they inevitably throw away information compared to the raw reading data.\looseness=-1

Moreover, the manner in which scanpaths are aggregated also impacts the empirical support for a given theory. 
E.g., surprisal theory \citep{hale-2001-probabilistic, levy2008expectation}, the theory that the processing effort for a linguistic unit depends on its in-context information content, suggests that contextual word predictability should have a more pronounced effect on gaze duration than total fixation time, because the latter can be influenced by material from the right context through regressive saccades.
But, counterintuitively, next-word surprisal has been found empirically to be a stronger predictor of total fixation time than gaze duration \citep{wilcox-2023-testing}.
Because both gaze and total duration are aggregate measures, providing a precise explanation for such results is challenging.
Another theoretical concern with aggregations is that they tend to conflate cognitive and oculomotor control processes. 
While cognitive processes, e.g., lexical access, contribute to reading slowdowns \citep{mollica2017incremental}, the oculomotor system also imposes physiological time delays between successive saccades, suggesting that reading times may reflect an interplay of both cognitive and mechanical constraints \citep{salthouse-ellis-1980-determinants,rayner-etal-1983-latency}.

In this paper, we advocate a unified approach that jointly models \emph{when} fixations occur, \emph{where} they land, and \emph{how long} they last. 
We achieve this by constructing a marked spatio-temporal point process, which alternates between generating saccades and fixations.
To model saccade timing and fixation locations (\emph{when} and \emph{where}), we employ a spatio-temporal \citet{hawkes1971} process, which captures how the density of future fixations changes in response to preceding ones in both time and space; \cref{fig:fixation_intensity} gives an illustration. 
To model fixation durations (\emph{how long}), we adopt a log-normal distribution with a convolution-based approach inspired by \citet{shain-schuler-2018-deconvolutional,shain-schuler-2020-continuous}.
We evaluate our probabilistic model on a re-processed (\Cref{sec:experimental-setup}) version of the English portion of the MECO corpus \citep{siegelman2022expanding}, assessing its ability to jointly model spatio-temporal disaggregated fixation and duration patterns.\looseness=-1

Empirically, we uncover several factors that yield better models of saccade planning, including, e.g., spatio-temporal dependencies on previous saccades, the ability to model the tendency towards a left-to-right progression of eye movement, and reader-specific effects.
Interestingly, these improvements are the most prevalent when fixations that occur outside of a character's bounding box are included in modeling.
When additionally including predictors like contextual surprisal, however, we observe only tiny improvements in the model's ability to perform saccade planning.
Furthermore, in our experiments on fixation durations, we obtain results that may have negative implications for established theories on human language processing. 
First, we observe little improvement when incorporating the effect of unboundedly many previous fixations through the convolution-based technique in comparison to a model that employs a Markov assumption when modeling past fixations.
This finding suggests that the effect of previous fixations on subsequent ones may indeed be bounded.
Second, when incorporating predictors such as contextual surprisal, unigram surprisal, and length, we obtain effects that are an order of magnitude smaller when modeling scanpaths as compared to modeling aggregated measurements. 
This finding suggests that some effects reported in the literature based on aggregated scanpaths may appear larger than they would if raw scanpath data were modeled directly.

\section{Modeling Reading Data}
\label{sec:reading}

While reading, our eyes make progress through the text via brief, rapid movements called \defn{saccades}.
Very little visual information is extracted during any one saccade \citep{IshidaIkeda89}. 
Instead, most of the information is extracted during the pauses that occur between saccades, where the eyes remain (mostly) stationary.\footnote{The eyes do not remain \emph{completely} stationary; there are constantly minor movements in the form of tremors, small drifts, and microsaccades \citep{rayner1998eye}.} These pauses, which are generally much longer than saccades, are called \defn{fixations}.
In these terms, reading can be thought of as alternating between fixations and saccades.

Examining reading behavior is one method to better understand the cognitive mechanisms that underlie reading and language processing more generally.
For example, fixations are known to reflect lexical access \citep{lima-inhoff-1985-lexical}, syntactic parsing \cite{frazier-rayner-1982-making}, and semantic integration \citep{ehrlich-rayner-1983-pronoun}.
A common way to measure reading behavior is through eye-tracking studies \cite{rayner1998eye}, which record high-frequency gaze samples that are segmented into discrete fixations. 

\paragraph{Formalization.}
Formally, each fixation is characterized as a triple $(\timestamp, \location, \duration)$ consisting of an \defn{onset time} $\timestamp$, a \defn{spatial location} $\location$, and a \defn{duration} $\duration$. The onset time $\timestamp \in \R_{\geq 0}$ is the starting time of the fixation relative to some reference point, typically the start of the recording. The spatial position lives in a bounded two-dimensional coordinate space $\screen \subset \R^2$, for example, a screen. 
Finally, the duration $\duration \in \R_{> 0}$ captures how long the eye remains still before initiating the next saccade.
We consider a (\defn{full}) \defn{scanpath} $\sequence$ to be a set of $N$ fixations, i.e., 
\begin{equation}
  \label{eq:fixation_sequence}
  \sequence \defeq \{(\ntime,\location_n,\ndur)\}_{n=1}^N,
\end{equation}
where $\mtime < \ntime$ if $m<n$.
For each $n \in \{1, \ldots, N\}$, we define the history $\nhist$ as
\begin{equation}\label{eq:fixation_events}
  \nhist \defeq
  \{(\mtime,\location_m,\mdur) \mid  \mtime \le \ntime \}.
\end{equation}
Note that in addition to fixations' onset times, locations, and durations, this sequence encodes the onset times and durations of saccades as well.\footnote{In particular, the onset of the $n^{\text{th}}$ saccade can be inferred by adding the $n^{\text{th}}$ duration to the $n^{\text{th}}$ fixation onset. The duration of the saccades can be inferred by taking the difference between the fixation onsets and the saccade onsets.} 
Furthermore, we note that \cref{eq:fixation_sequence} includes the full sequence of fixations and saccades that occur in $\screen$ during an eye-tracking session. This includes fixations that do not land on any word, i.e., those that land outside character bounding boxes, re-fixations on different parts of the same word, and regressive saccades. We also consider \defn{filtered scanpaths}, i.e., the subsequence of a full scanpath that only contains fixations that fall within the bounding box of some word.\footnote{Fixations outside word bounding boxes account for 52.3\% of all fixations in our dataset. Note that both the size of a bounding box and the strategy for assigning fixations to words are determined at the discretion of the modeler.}\looseness=-1

\subsection{Modeling Details}\label{sec:previous-work-aggregations}

\paragraph{Aggregations.} 
In most sentence processing experiments based on eye-tracking data, the raw data are typically aggregated into reading time variables at the word level; see \citep{frank-etal-2013-reading}.\footnote{Aggregations into other regions of interest are also used \citep{Cook2019-hm,brodbeck2022parallel}---especially when studying specific types of reading behavior like skip rates \citep{rayner2011eye,giulianelli-etal-2024-proper}.}
Some of the most common word-level aggregations are:\looseness=-1
\begin{itemize}[noitemsep,topsep=0.3em,leftmargin=*]
\item \defn{first fixation duration}, the duration of the first fixation that lands on a word;
\item \defn{gaze duration}, the sum of the durations of all fixations that land on a word before leaving it the first time;
\item \defn{total fixation duration}, the summed duration of all the fixations on a word; and
\item \defn{scanpath duration}, the summed duration of all \emph{consecutive} fixations that land on the same word. \looseness=-1
\end{itemize}
 Note that the first three aggregation strategies are ordered in a nested fashion, i.e., each strategy aggregates over at least as many fixations as the previous; for more details, see \citet{inhoff1984stages}, \citet{berzak-2023-eye}, and \Cref{appendix:aggregations-review}.
The final aggregation strategy, scanpath duration, is \emph{not} nested in the sense above.
Moreover, a sequence of scanpath durations may contain multiple durations that correspond to the same word.
See \citet{shain-schuler-2020-continuous} for more details. 

\paragraph{Statistical analysis.} 
The standard approach to analyzing aggregated measurements is to use linear modeling \citep{kliegl-2007-toward,giulianelli-etal-2024-generalized,kuribayashi-etal-2024-psychometric,kuribayashi2025largelanguagemodelshumanlike} or generalized additive modeling \citep[GAMs;][]{smith2013-log-reading-time,goodkind-bicknell-2018-predictive,wilcox2020,wilcox-2023-testing,klein-etal-2024-effect}.
When multiple reading times per stimulus are available, e.g., when multiple participants read the same stimulus in their respective trials, it is appropriate to apply a mixed-effects model \citep[e.g.,][]{aurnhammer-frank-2019,xu-etal-2023-linearity}, using random effects to account for variability across participants. 
In \cref{sec:experiments-results}, we investigate whether incorporating reader-specific effects improves the modeling of saccades and fixation durations, estimating them as fixed effects rather than random effects.

\paragraph{Spillover effects.} Beyond aggregation, there is another limitation of the above-mentioned modeling techniques worth noting. The effect that cognitive processing of a unit has on reading time is not necessarily instantaneous; it often leads to reading slowdowns for units that occur later in the text \citep{shain-schuler-2020-continuous}.
To incorporate such effects, these models must include additional predictors describing previous units, which are called \defn{spillover variables}. 
We discuss the modeling of spillover variables in more detail in \cref{sec:duration-model}.

\paragraph{Predictors.} It is common to use predictors derived from pre-trained language models when modeling reading data in order to estimate the effect that the prior context has on reading time \citep[e.g.,][]{oh2023does}.
Here, we focus on one such predictor called \defn{contextual surprisal}.
Let $\alphabet$ be a finite, non-empty set of linguistic units, e.g., characters or words, called an \defn{alphabet}, and let $\alphabet^*$ be the set of all strings that can be formed by concatenating units in $\alphabet$, including the empty string $\varepsilon$.
We further employ a special symbol $\eos \notin \alphabet$ to denote the end of a string, and define $\eosalphabet \defeq \alphabet \cup \{\eos\}$.
Following \citeposs{shannon1948mathematical} formulation of information content, 
the \defn{contextual surprisal} of a unit \(\unit_{\timestamp} \in \eosalphabet\) 
given a preceding context of units $\ctx \in \alphabet^*$ is defined as\looseness=-1
\begin{equation}\label{eq:surprisal}
    s_{\timestamp}(\unit_{\timestamp}) \defeq -\log_{2}\,\overrightarrow{p}(\unit_{\timestamp} \mid \ctx),
\end{equation}
where $\overrightarrow{p}(\cdot \mid \ctx)$ is the true (albeit unknown) distribution over $\eosalphabet$ conditioned on the context \(\ctx\), defined using prefix probabilities; see, e.g., \citealp{opedal-etal-2024-role}, for further technical discussion. 

\paragraph{Surprisal theory.}
Surprisal theory \citep{hale-2001-probabilistic, levy2008expectation} predicts that the cognitive processing effort for a linguistic unit $\unit$ is a function of its contextual surprisal.
This theory has received empirical support across several datasets and languages \citep{kuribayashi-etal-2021-lower,wilcox-2023-testing}, and there is evidence suggesting that the linking function between surprisal and different aggregated measurements of processing effort is well-approximated by an affine function \citep{smith2013-log-reading-time,xu-etal-2023-linearity,shainetal24}. 
Such studies typically include additional baseline predictors, like unigram frequency and word length \citep{opedal-etal-2024-role}.
Importantly, surprisal theory is a theory about computational demand, situated at the computational level of \citeposs{marr1982} hierarchy. 
Thus, it neither makes direct predictions about the underlying cognitive mechanisms nor about eye movement control.\footnote{For more discussion, see \citet{ohams_nair_bhattasali_resnik_2025} who propose a mechanistic model of human sentence processing based on predictive coding \citep{Friston2009-sj}.} 
In our empirical study presented in \cref{sec:saccade-experiments}, we explore whether surprisal and other predictors are predictive of more fine-grained eye movements than the theory itself considers, i.e., whether they are useful for \defn{saccade planning} \citep{ZINGALE19871327}. \cref{sec:related-models} gives a brief overview of related cognitive models of reading.\looseness=-1

\section{Modeling Fine-Grained Reading Data}
\label{sec:model}

To capture the fine-grained dynamics of human reading, we now propose a probabilistic model of scanpaths.
At the core of our approach is the goal of generating the three key components of a fixation: \emph{when} it begins, \emph{where} it lands on the screen, and \emph{how long} it lasts. 
Our probabilistic model takes the form of a spatio-temporal \emph{marked} point process, consisting of two components.
\begin{itemize}[leftmargin=1.5em]
    \item A \textbf{spatio-temporal point process} for fixation onsets and locations, characterized by a density function $f(\ntime, \nloc \mid \nphist )$ that models the likelihood that the $n^{\text{th}}$ fixation in the scanpath $\sequence$ occurs at time $\ntime$ on screen location $\nloc$ given the history of preceding fixations up to time $\nptime$. 
    See \cref{sec:hawkes-process} for details.
    \item A \textbf{probability distribution} for (non-negative) fixation durations (\defn{marks}), which is characterized by a density function $\durdensity (\ndur \mid \nphist, \ntime)$ that models the likelihood of the $n^{\text{th}}$ fixation in the scanpath having duration $\ndur$ given the last fixation onset $\ntime$ and the history of preceding fixations up to time $\nptime$. 
    See \cref{sec:duration-model} for details.\looseness=-1
\end{itemize}
Under this model, a scanpath is iteratively sampled as follows.
Note that $\history_0 \defeq \{\}$.
\begin{enumerate}[label=(\alph*), leftmargin=1.5em]
    \item \textbf{Sample a fixation's onset time and location}: 
    \begin{equation} 
    (\ntime, \nloc) \sim \hawksdensity(\cdot, \cdot \mid \nphist),
    \end{equation}
    according to the spatio-temporal point process; see \Cref{eq:hawkesdensity-ours} for more details.

    \item \textbf{Sample a fixation's duration}:
    \begin{equation}
    \ndur \sim \durdensity(\cdot \mid\nphist, \ntime),
    \end{equation}
    according to the density of fixation durations; see \Cref{eq:duration} for more details.\footnote{The density $\durdensity(\cdot \mid\nphist, \ntime)$ could also have been conditioned on the $n^{\text{th}}$ screen location $\nloc$; we chose not to do so based on results obtained from preliminary experiments.}

    \item \textbf{Update the history with new fixation}: The full fixation $ (\ntime, \nloc, \ndur)$ is added to the history:
    \begin{equation}
    \nhist \defeq \nphist \cup \{ (\ntime, \nloc, \ndur ) \}.
    \label{eq:history_update}
\end{equation}
\end{enumerate}
Steps~(a)–(c) are iterated until a predefined reading horizon $T \in \R_{\geq 0}$ is reached.\footnote{An application of our model that we do not emphasize here is generating synthetic scanpaths. Besides being interesting for cognitive modeling, synthetic scanpaths have been used both to enhance language representations \citep{barrett-etal-2018-unsupervised,sood2020improving} and to probe the internal behavior of neural NLP models \citep{sood-etal-2020-interpreting,hollenstein-etal-2022-patterns}.\looseness=-1}
Thus, our model can be viewed as a \defn{stopped} spatio-temporal marked point process \cite{daley1988introduction}.\looseness=-1

\subsection{Modeling Fixation Onsets and Locations}
\label{sec:hawkes-process}

We model $\hawksdensity(\ntime, \nloc\!\mid\!\nphist)$ with a spatio-temporal \defn{Hawkes} process \cite{hawkes1971}.\footnote{Spatio-temporal Hawkes processes \citep{reinhart2018review} have several other applications, e.g., modeling earthquakes and their aftershocks \citep{Ogata1999}, urban crime \citep{Mohler2011crime}, and spread of infectious diseases \citep{Meyer2011-tr}.} 
A Hawkes process is a special case of a spatio-temporal non-homogeneous Poisson point process, which assumes that the probability that a new event occurs in the (infinitesimal) region $[\ntime, \ntime+\mathrm{d}\ntime) \times [\nloc, \nloc+\mathrm{d}\nloc)$ is approximately \(\intensity(\ntime,\nloc ; \nphist)\mathrm{d}\ntime \mathrm{d}\nloc\), where $\intensity\colon \R_{\geq 0} \times \screen \rightarrow \R_{\geq 0}$ is called the \defn{intensity function}. 
A Hawkes process allows past fixations to influence the probability of new fixations in an additive manner.
This influence is often termed \textit{self-excitation} in the literature.
The intensity function of a spatio-temporal Hawkes process is defined as
\begin{align} \label{eq:intensity-hawks}
\intensity(&\ntime, \nloc ; \nphist)
  \defeq  \\
  &  \baseintensity + \sum_{m = 1}^{n-1}
      \tempkern_{m}\left( \ntime - \mtime -\adjterm(n,m) \right)\spatdens_m(\nloc),   \nonumber 
\end{align}
where $\baseintensity \in \R_{\geq0}$ is the base intensity, $\tempkern_m\colon \R_{\geq 0}\rightarrow \R_{\geq 0}$ are exponentially decaying temporal kernels governing the influence of past events, $\spatdens_m\colon \Omega \rightarrow \R_{\geq 0}$ is a density over the coordinate space $\screen$, and 
\begin{align}
\adjterm(n,m)
  &\defeq \sum_{j =1}^{n-1} d_j-
     \sum_{j=1}^{m-1} d_j 
\end{align}
is the cumulative duration between $\ntime$ and $\mtime$.
Notationally, we suppress the dependence of $   \tempkern_{m}$, $\spatdens_m$, and $\adjterm$ on the history $ \nphist$ for simplicity. 
Subtracting $\adjterm(n,m)$ from the time difference in \Cref{eq:intensity-hawks} ensures that the kernel $\tempkern_m(\cdot)$ only captures the intervals between saccades.
Furthermore, we remark that, when generating the $n^{\text{th}}$ fixation, the Hawkes process conditions on \emph{all} past fixations $\nphist$, which is what enables the self-exciting behavior.\looseness=-1

\paragraph{Temporal kernel.}
Our choice of $\tempkern_m$ allows each past fixation indexed by $m$ to contribute to the intensity of future fixations. 
However, that influence decays exponentially the farther back it looks, as determined by $\ntime - \mtime -\adjterm(n, m)$ from \cref{eq:intensity-hawks}.
We consider an exponential kernel, parameterized as\looseness=-1
\begin{equation}    
  \tempkern_m(\Delta)
  \defeq
  \tempact( \tpredictors \alphapar )
  \exp(-\tempact(\tpredictors\betapar) \cdot \Delta),
\label{eq:exp_dec}
\end{equation}
where $\predictors \in \R^p$ is a column vector of predictor values associated with the $m^{\text{th}}$ fixation (e.g., surprisal), $\alphapar$ and $\betapar$ are column vectors in $\R^p$ of learnable parameters, and $\tempact \colon \R \rightarrow \R_{\geq 0}$ is a function that ensures the non-negativity of the dot product, e.g., a ReLU.
Note that $\tempact( \tpredictors \alphapar)$ quantifies how much a fixation increases the probability of subsequent fixations, its excitation strength, while $\tempact(\tpredictors \betapar)$ determines how quickly the influence of the fixation diminishes over time, its decay rate.
Because excitation strength and decay rate depend on an event-specific vector of predictor values $\predictors$, the strength of excitation and decay rate may vary across different spatio-temporal conditions.\looseness=-1

\paragraph{Spatial density.}
We model the spatial component $\spatdens_m$ as a bivariate spherical Gaussian distribution centered at $\spatmean (\mloc)$, where $\spatmean \colon \R^2 \rightarrow \R^2$ is a fixation-specific transformation function; we describe specific choices of $\spatmean$ in the paragraph below.
These choices result in the following density
\begin{align}
  \spatdens_m(\nloc)
  \defeq
  \frac{1}{2\pi\sigma^2}
  \exp\left(
  -\tfrac{\|\nloc - \spatmean(\mloc)\|^2}{2\,\sigma^2}
  \right),
\label{eq:spatial_distr}
\end{align}
where $\|\cdot\|$ denotes the Euclidean ($L_2$) norm.
This design choice allows us to interpret the intensity function introduced in \cref{eq:intensity-hawks} as a conical combination of Gaussian densities, each corresponding to a past fixation event.
Thus, the intensity function can be seen as proportional to the density of a multi-modal distribution over $\screen$.
We contend that this property aligns well with what is known about human reading behavior because it captures that eye fixations may jump either forward or backward in the text with varying probabilities \citep{rayner1998eye}.

We consider three different choices of the transformation function $\spatmean$.
As a \underline{b}aseline, we first consider $\spatmeanbase(\location) \defeq \location$, i.e., the identity function. However, this baseline yields an overly simple model: (i) it predicts that saccades are most likely to occur around previous fixations, and (ii) it does not incorporate other predictors that may influence the location of the next fixation. 
To address (i), we parameterize $\spatmean$ as an affine transformation of the previous fixation location 
$\location$.
Mathematically, this means that $\spatmean(\location)$ takes the following \underline{a}ffine form 
\begin{equation}\label{eq:spatial-mean-affine}
    \spatmeanaff(\location) \defeq \spatialparamslocation\location + \spatialparamsbias,
\end{equation} 
where $\spatialparamslocation \in \R^{2\times2}$ is a matrix and $\spatialparamsbias \in \R^2$ is a vector.
To address (ii), we consider the same vector of predictors $\predictors \in \R^p$ applied into the temporal kernel and incorporate predictor-specific effects, yielding the \underline{f}ully parameterized transformation function
\begin{equation}\label{eq:spatial-mean-extended}
    \spatmeanfull(\location) \defeq  \spatmeanaff(\location) + \spatialparamspreds \predictors,
\end{equation} 
 where $\spatialparamspreds \in \R^{2\times p}$ is a matrix.

\paragraph{The density function.}
Given the intensity function defined in \Cref{eq:intensity-hawks}, the corresponding probability density function of a Hawkes process is given by\looseness=-1
\begin{align}\label{eq:hawkesdensity-ours}
  \hawksdensity(&\ntime, \nloc \mid \nphist)
\\
&\defeq \frac{
    \intensity(\ntime, \nloc ; \nphist)
  } 
  {\exp(\bigint(\ntime ; \nphist))} \mathbbm{1}\{\ntime \geq \nptime + \npdur\}. \nonumber
   \nonumber
\end{align}
The indicator function ensures that no probability mass is distributed to preceding time points and
\begin{align}
\bigint(\ntime ; &\,\nphist) \\
&\defeq
      \int_{\nptime + \npdur}^{\ntime}
        \int_{\Omega}
          \intensity(u,\bm{z}; \nphist)\dinf \bm{z}\dinf u. \nonumber
\end{align}
See \citet[\S7.2]{daley1988introduction} for a detailed technical discussion.

\subsection{Modeling Fixation Durations}\label{sec:duration-model}

Fixation durations are modeled as non-negative random variables with a conditional density denoted by $\durdensity(\ndur \mid \nphist, \ntime)$.
Let $\ndur > 0$ be the duration of the $n^{\text{th}}$ fixation. 
We assume a log-normal distribution for fixation durations,\footnote{We evaluated several candidate distributions using $K$-fold cross-validation and found that the log-normal distribution offers a good trade-off between interpretability and goodness-of-fit; it outperformed alternatives such as the Weibull, exponential, and Rayleigh distributions. See \cref{appendix:lognormal} for details.}
i.e.,
\begin{align}\label{eq:duration}
\durdensity(\ndur \mid & \,\nphist, \ntime)  \defeq \\
  & 
\frac{1}{\ndur \sqrt{2\pi \sigma^2}} \exp\left(
- \frac{(\log \ndur - \durmean(\ntime))^2}{2\sigma^2}
\right), \nonumber
\end{align} 
where $\durmean(\ntime)$ is a fixation-specific function of the fixation onset $\ntime$ and the past fixation history $\nphist$. 
For clarity, again, we suppress the explicit dependence of $\durmean(\ntime)$ on $\nphist$ in the notation. 
Note that $\sigma^2 > 0$ is a fixed variance parameter, different than $\sigma^2$ in \cref{eq:spatial_distr}.\looseness=-1

\paragraph{Conditional log-mean of duration.} 
Similar to \citet{shain-schuler-2018-deconvolutional, shain-schuler-2020-continuous}, we model temporally delayed spillover effects through a convolution over past predictor values. Let  $\predictors \in \R^p$ be a column vector of predictor values corresponding to the $m^{\text{th}}$ fixation and $K \subseteq \{1, \dots, p\}$ be a set of indices representing the predictors for which we model spillover effects. 
We then define the \underline{c}onvolution function \looseness=-1
\begin{align}
    \label{eq:log_mean_duration}
 \durmeanconv(&\ntime) \defeq  \tdurpredictors \weights \\ & +   \sum_{k \in K} w^\prime_k \sum_{m=1}^{n-1} x_{mk} \gammadis \left(\ntime - \mtime \mid \agamma, \bgamma, \thgamma \right) \nonumber,
\end{align}
where $\weights \in \R^p$ is a column vector of parameters, $x_{mk}$ denotes the value of predictor $k$ at fixation $m$, $w^\prime_k \in \R$ is the spillover coefficient of predictor $k$, and the kernel $\gammadis(\cdot \mid \agamma, \bgamma, \thgamma)$ is the density function of a shifted gamma distribution, defined as 
    \begin{align}
        \gammadis(\tau \mid & \,\agamma, \bgamma,\thgamma)
  \defeq \\
  &\frac{\bgamma^{\agamma}}{\Gamma(\agamma)}
    (\tau+\thgamma)^{\agamma-1}
    \exp\left(-\bgamma(\tau+\thgamma)\right), \nonumber
    \end{align}
with parameters $\agamma \in \R_{>1}$, $\bgamma \in \R_{> 0}$ and $\thgamma \in \R_{\geq 0}$; $\Gamma(\cdot)$ denotes the gamma function.

\paragraph{Markovian spillovers.}
Most previous work (see \cref{sec:previous-work-aggregations}) incorporates spillover effects as additional predictors in a linear model or GAM.
In our framework, this corresponds to replacing the gamma kernel in \cref{eq:log_mean_duration} with spillover parameters $\spilledcoeffs \in \R$, yielding the \underline{M}arkov model:
\begin{align}
\label{eq:log_mean_duration_spillover_variables}
 \durmeanspill(\ntime) \defeq  \tdurpredictors \weights + \sum_{k \in K} \sum_{m=n-\paststeps}^{n-1} \spilledpred \spilledcoeffs,
\end{align}
where $\paststeps$ denotes the number of previous fixations that contribute to the spillover effect, i.e., the model's Markov order.
Each predictor $\spilledpred$ represents the value of feature $k \in K$ at time step $m$, and $\spilledcoeffs \in \mathbb{R}$ is the corresponding weight applied to that feature and lag. The double subscript emphasizes that the model learns distinct weights for each combination of feature and lag.

\section{Experiments and Results}\label{sec:experiments-results}

We employ the modeling framework introduced in \cref{sec:model} to run two suites of experiments distinguished by the scanpath components they target, i.e., saccade planning (\cref{sec:saccade-experiments}) and fixation durations (\cref{sec:duration-model-experiments}). The experimental setup is detailed in \Cref{sec:experimental-setup}.
In each of the subsequent subsections, we introduce the hypotheses we are testing, the model specifications we use, and present experimental results.

\begin{figure*}
    \centering
    \includegraphics[width=\textwidth]{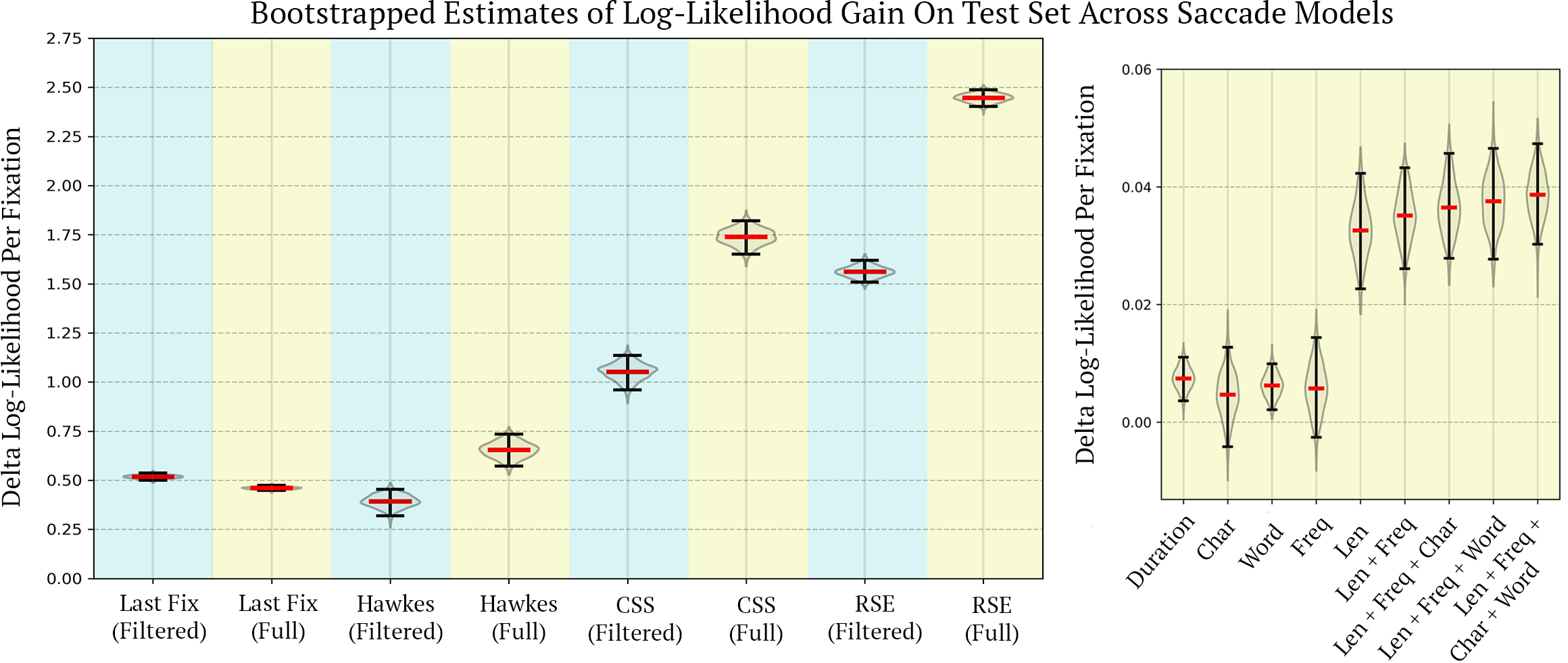}
    \vspace{-16pt}
    \caption{
Bootstrapped distributions of per-fixation log-likelihood gains for different saccade models, evaluated relative to the Poisson baseline (left) and the RSE model (right). Higher values indicate better predictive performance in modeling saccade behavior. The models in the left-hand panel are described in \cref{sec:saccade-experiments}, while the models in the right panel include the predictors introduced in \cref{sec:experimental-setup}. We distinguish between whether the model was trained and evaluated on the full or the filtered scanpaths. 
The right-hand panel reports results only for the full dataset; an extended version is given in \cref{fig:hawkes_process_results_extended} (\cref{sec:more-results}).
For the predictors, we write ``word'' for word-level surprisal, ``char'' for character-level surprisal, and ``freq'' for unigram surprisal. 
    }
    \label{fig:hawkes_process_results}
\vspace{-8pt}
\end{figure*}

\subsection{Modeling Saccade Planning
}\label{sec:saccade-experiments}

In this section, we empirically study how well the spatio-temporal Hawkes process fits to saccades in reader scanpaths. We are interested in a number of different questions.
 First, we ask which components of the model are useful for modeling saccades. To this end, we introduce several model specifications, including a learned constant spatial shift and reader-specific effects. We also introduce two simple baseline models. 
Second, we are interested in whether saccades can be modeled using filtered scanpaths, or whether this requires more fine-grained data.
 Finally, we turn our attention towards the predictors. We investigate whether incorporating word and character-level quantities as predictors associated with past fixations helps determine where and when the next saccade will land, beyond what is captured by constant displacements from past fixations and individual reader effects.

\subsubsection{Model Specifications}\label{sec:specifications-saccade}

\paragraph{Baseline models.} We obtain baseline models by simplifying the intensity function from \cref{eq:intensity-hawks}.
The main baseline model is a \defn{standard Hawkes process}. It defines the excitation strength $\alpha$ and decay rate $\beta$ in \cref{eq:exp_dec} as scalar values, and $\spatmean(\cdot)$ from \cref{eq:spatial_distr} as the identity function. We also include two simpler models: a \defn{last-fixation} baseline which models each fixation as normally distributed around the previous one with constant variance, and a \defn{Poisson} baseline which assumes fixations to be uniformly and independently distributed in space and time. Further details are given in \cref{app:baseline-models}.

\paragraph{More sophisticated models.} 
Next, we introduce models that are more expressive than our baselines. 
Our first model is the \defn{\underline{c}onstant \underline{s}patial \underline{s}hift} (CSS) model, which predicts new fixation locations based on a constant displacement from previous fixations. 
This term captures a baseline tendency for gaze to progress linearly across the text, independent of specific word content or surrounding linguistic context. It uses the spatial mean from \cref{eq:spatial-mean-affine}. 
This model and our baseline models treat all scanpaths as if a single average reader generated them; however, this treatment deviates from the reality that all individuals exhibit distinct reading styles. 
To capture such variations, we include reader-specific fixation effects in both the temporal and spatial parameters for each of the $R \in \mathbb{Z}_{>0}$ readers.
We consider a vector of predictors 
\begin{align}
    \label{eq:readers-pred}
    \predictors \defeq \mathbf{1} \oplus \mathbf{u}_m,
\end{align}
 where $\mathbf{1} \in \R^1$ is the $1$-dimensional column vector with value $1$ corresponding to the intercept, $\mathbf{u}_m \in \{0,1\}^{R}$ is a unit vector with a single non-zero value that corresponds to a particular reader; $\oplus$ denotes column-wise concatenation.
As a result, the dot products $\tpredictors \alphapar$ and $\tpredictors \betapar$ will be the sum of a global effect and a reader-specific effect.
Note that this influences both the temporal \pcref{eq:exp_dec} and the spatial kernel \pcref{eq:spatial-mean-extended}. We refer to this model as the \defn{\underline{r}eader-\underline{s}pecific \underline{e}ffects} (RSE) model.
Finally, we treat each predictor associated with a fixation as an effect that modulates both the temporal kernel and the spatial mean. In the case of a single effect $z$, we let the predictor vector be equal to \looseness=-1
\begin{equation}\label{eq:predictors-full-hawkes}
    \predictors \defeq \mathbf{1} \oplus \mathbf{u}_m \oplus z_m\mathbf{1} \oplus z_m\mathbf{u}_m.
\end{equation}
Here, $z_m \mathbf{1}$ captures a global effect and $z_m \mathbf{u}_m$ captures its interaction with the reader identity encoded in $\mathbf{u}_m$.
For saccades that do not land on a word and therefore lack a corresponding effect value, we include binary indicators denoting the presence of the effect, along with their reader interactions, in \Cref{eq:spatial-mean-extended}.

\subsubsection{Results}

\Cref{fig:hawkes_process_results} shows bootstrapped estimates of per-fixation log-likelihood improvements for the various saccade models evaluated on the held-out test set, as specified in \cref{sec:experimental-setup}.
In the left panel, we show the results when using the Poisson process as a baseline.
The results demonstrate that the increasingly expressive models specified above, i.e., those resulting from the incorporation of temporal dependencies, the constant spatial shift, and reader-specific modeling, help the model to better generalize to held-out data.
The best-performing model in the left panel, obtained by inserting \cref{eq:readers-pred} into \cref{eq:exp_dec} and \cref{eq:spatial-mean-extended}, achieves an average per-fixation log-likelihood gain of 2.44 nats over the Poisson baseline.
This corresponds to the model assigning approximately $(\exp(2.44) -1) \times 100\approx 1047 \%$ higher likelihood to the true next fixation on average.
Parameter estimates from this model suggest a consistent global rightward shift of roughly $\approx 10.61$ characters (including white spaces) further to the right, a pattern consistent with the left-to-right progression of (English) writing. 
There is also evidence of self-excitation.
Given estimates $\alphahat$ and $\betahat$ of the parameters in \cref{eq:exp_dec}, we find $\alphaest = 12.64 \pm 2.69$ and $\betaest = 16.24 \pm 3.44$, with values corresponding to the mean and standard deviation across individual readers, and $\predictors$ as defined in \cref{eq:readers-pred}.
However, the lower magnitude of $\alphaest$ in comparison to $\betaest$ suggests that, although recent fixations do influence the likelihood of another fixation in short succession, this influence decays relatively quickly over time.

The right panel of \Cref{fig:hawkes_process_results} shows the marginal effect of adding lexical predictors on top of the RSE model. While improvements are statistically significant, they are modest in magnitude: most predictors yield under 2\% relative gain in per-fixation log-likelihood, a much smaller increase than what can be observed in the left panel. Word length displays the highest predictive power, leading to performance improvements by approximately 4\%. This may be due to the fact that a longer word, in general, requires a longer saccade to move to future words.\looseness=-1 

Finally, in \cref{fig:hawkes_process_results}, we observe that models trained on full scanpaths place a higher probability on held-out data than those trained on filtered scanpaths.
This pattern suggests that it is useful to preserve the full sequence of preceding fixations, rather than only those associated with words. 
See \cref{fig:hawkes_process_results_extended} for an extended comparison.\looseness=-1

\subsection{Modeling Fixation Durations}\label{sec:duration-model-experiments}
In this section, we focus on modeling fixation durations through the log-normal, convolution-based model given in \cref{eq:log_mean_duration}, comparing it to the linear Markov model given in \cref{eq:log_mean_duration_spillover_variables}. In addition to modeling the full scanpaths, we also fit linear models to the three aggregated datasets described in \cref{sec:experimental-setup}. We are interested in whether surprisal and other predictors are effective across several levels of aggregation, or whether the empirical support for, e.g., surprisal theory, depends on the aggregation strategy. For the linear models, we log-transform the response variable. \looseness=-1

\subsubsection{Model Specifications}
\label{sec:duration-model-experiments:model-specs}
Similarly to in \cref{sec:specifications-saccade}, we introduce reader-specific effects and the predictors described in \cref{sec:experimental-setup} into the predictor vector $\durpredictors$. Specifically, we define $\durpredictors$ analogously to \cref{eq:readers-pred,eq:predictors-full-hawkes}, respectively, where each predictor is associated with the $n^{\text{th}}$ fixation. Each predictor vector is then used in \cref{eq:log_mean_duration} and \cref{eq:log_mean_duration_spillover_variables}.
When applying \cref{eq:log_mean_duration_spillover_variables}, we set the number of previous fixations $l=2$.
The baseline models contain intercepts, including reader-specific ones when modeling non-aggregated data, as well as durations of the fixations in the history. The latter are included as controls in order not to attribute the effects of past durations to the predictors. We also perform experiments in which the baseline \emph{excludes} the prior fixation durations to get a sense of how much the effect sizes reduce. 

\subsubsection{Results}
\cref{fig:duration_convolution_results_figure} (in \cref{sec:more-results}) plots bootstrapped distributions of per-fixation log-likelihood improvements for models that incorporate spillover effects via a convolution term, evaluated on the full and filtered scanpath datasets. 
Note that these are the only datasets that retain fixation-level timing. We observe very small and mostly insignificant effect sizes: With the exception of the model combining length, unigram suprisal, and character-level surprisal, none of the intervals exclude zero, and all average effect sizes remain below~$0.004$. Comparing the same models on the full versus filtered scanpaths, we generally observe slightly larger average improvements for the filtered scanpaths. 
This raises the question of whether different data aggregation strategies could recover the larger effect sizes previously reported (e.g., \citealp{wilcox-2023-testing}).\looseness=-1

In \Cref{fig:linear_model_fix_duration_results} (\cref{sec:more-results}), we present results for the Markov linear models introduced in \cref{eq:log_mean_duration_spillover_variables} across differently aggregated datasets and across predictors. 
First, comparing the log-likelihood improvements with those in \cref{fig:duration_convolution_results_figure} shows that the convolution-based model and the linear Markov model achieve similar fit.
Notably, the convolution-based model, despite its greater expressiveness, does not yield larger improvements than the linear model. These results suggest that it is effective in practice to model fixation durations with a linear model and Markov assumption on the spillover effects, at least on the MECO dataset.

Comparing across aggregation schemes, we find that averaging gaze durations across readers leads to substantially larger mean log-likelihood improvements across folds, ranging from approximately $0.023$ to $0.037$ when the baseline includes duration spillovers (see \Cref{fig:linear_model_fix_duration_results}); this is indeed consistent with the effect sizes reported in Fig. 1 from \citet{wilcox-2023-testing}.
In contrast, disaggregated (reader-specific) models yield improvements that are more than an order of magnitude smaller, with effect sizes of up to only $0.003$ relative to the same spillover baselines.
Aggregation also introduces greater variance, with the effect sizes in the aggregated setting ranging from $0.01$ to $0.05$ across folds, when accounting for duration spillovers.
As for the filtered and full scanpaths, the effect sizes reduce even further. 
Furthermore, we find that effects are consistently lower when past durations are included in the baseline model (cf.\ \Cref{sec:duration-model-experiments:model-specs}). We thus advocate for the continued use of such more stringent baselines in future studies.
Finally, we note that the absolute effects on aggregated data are somewhat smaller than what has been reported in previous studies on MECO data \citep[Figs. 4 and 5,][]{opedal-etal-2024-role}.
This may be due to differences in preprocessing: while we map fixations to words via bounding boxes (see \cref{sec:experimental-setup}), \citet{siegelman2022expanding} appear to have used a different method.\looseness=-1

\section{Conclusion}
We introduced a probabilistic model of reading behavior, through which we investigate modeling strategies for saccade planning and evaluate whether fine-grained models of individual fixation durations outperform approaches based on aggregated gaze measurements.
With respect to saccade planning, our findings highlight the importance of including the complete fixation history while also accounting for systematic rightward tendencies in gaze as well as reader-specific behavior. 
We also observe that surprisal-based predictors contribute minimal additional predictive power.
When modeling fixation durations, we similarly observe that incorporating surprisal-based predictors in unfiltered scanpath models only yields marginal improvements in predictive power. 
Taken together, our findings suggest that commonly used aggregation and data processing procedures can substantially influence the outcomes of reading time analyses.\looseness=-1

\section*{Limitations}

This study contributes to the literature on modeling techniques for psycholinguistic data, but several limitations warrant discussion.
First, in terms of data, our analysis is restricted to the English portion of MECO, leaving the other ten languages unexamined. Cross-linguistic validation is needed to assess whether our findings generalize to languages with distinct orthographic or syntactic properties (e.g., languages that are read from right to left). 
In addition, the relationship between the predictors and the responses was intentionally constrained to be affine, consistent with findings from previous work; see \cref{sec:previous-work-aggregations}. This may oversimplify the relationship between cognitive processes and eye movements compared to more expressive modeling choices.\looseness=-1 

With respect to modeling choices, another limitation---raised during the reviewer discussion---concerns our assumption of isotropic variance in the kernel for the spatial density function \pcref{eq:spatial_distr}. While this assumption was made out of simplicity, it may limit the model's ability to capture directional biases in eye movements. A natural extension would be to replace the isotropic kernel with an anisotropic one, or to apply non-linear warping functions to better accommodate complex spatial patterns in fixation behavior.\looseness=-1

One reviewer also raised the importance of capturing structured spatial effects, particularly those related to line transitions, and we explored one such direction. Specifically, we experimented with augmenting the model with a feature encoding the distance from the current fixation to the right margin, activated via a ReLU function to signal likely transitions to a new line. However, we encountered difficulties during training, as this feature failed to produce stable non-zero gradients, possibly due to the sparsity of fixations near line endings. For this reason, we chose not to include this model variant in the final set of results presented here.

Additional directions for extending the modeling framework can be drawn from existing literature. For instance, the framework could be extended to support neural Hawkes processes \citep{mei-eisner-2017}, including for the modeling of individual reader effects \citep{boyd2020userdependent}. Another interesting direction would be to study counterfactual reading scenarios---for example, asking questions about how scanpaths are influenced by external disturbances that may disrupt reading. There already exist techniques to estimate causal effects under temporal point processes \citep{gao2021causalmultivariatepoint,noorbakhsh2022counterfactual,zhang2022counterfactual}, which could be applied and extended to the setting considered here.\looseness-1

Regarding the experimental results, when training the PyTorch models, we conducted a grid search over a range of hyperparameters to identify the best-performing configurations. However, we cannot guarantee that these configurations correspond to globally optimal parameter estimates.
We also note that the character- and word-level surprisal estimates are derived from different language models.
Thus, any performance differences between them cannot be attributed solely to the granularity of representation, i.e., character- vs. word-level.
Additionally, the standard Hawkes process baseline and the last-fixation baseline exhibited comparable performance. While we extended the Hawkes model to capture spatial gaze patterns, we did not apply similar enhancements to the last-fixation baseline. A more balanced comparison---where both models are equipped with spatial effects---could offer clearer insights into their relative strengths.

\section*{Acknowledgments}
We thank the reviewers from ACL Rolling Review for their valuable feedback and discussion. AO was supported by the Max Planck ETH Center for Learning Systems.
MG was supported by an ETH Zürich Postdoctoral Fellowship. 
RC gratefully acknowledges support from the Hasler Foundation.

\bibliography{custom}

\onecolumn
\appendix

\section{Experimental Setup}
\label{sec:experimental-setup}

\paragraph{Data.} We use the English portion of the MECO dataset \citep{siegelman2022expanding} as a source of reading data. 
The MECO dataset provides scanpaths for multiple readers over 12 short text excerpts from Wikipedia; the English portion of the dataset includes scanpaths for 46 readers. The texts were displayed on a 1920$\times$1080 screen in monospaced Consolas 22pt font. As summarized in \Cref{tab:summary_bbox}, the number of characters per text ranges from 831 to 1230 (arithmetic mean 1093), and the number of lines from 8 to 12 (arithmetic mean 10.5).
We use the Tesseract OCR library\footnote{\url{https://pypi.org/project/pytesseract/}} in Python to identify characters and their bounding boxes; see the paragraph below for more details. 
For each session, we consolidated the gaze measurements into a scanpath sequence $\sequence$  for each reader. This process resulted in 97,742 fixations across texts and readers.
Each fixation is then associated with the character whose bounding box contains its location coordinates or flagged as having landed outside the bounding box of any character.
To facilitate comparison with coarser targets, we derived three additional datasets: (i) the filtered \defn{scanpath-duration dataset} (\cref{sec:reading}), obtained by removing fixations outside word bounding boxes and merging consecutive fixations on the same word (46,511 data points); (ii) the \defn{per-reader word-level gaze-duration dataset}, in which each record captures the gaze duration of a single reader on a single word (34,368 data points); and (iii) the \defn{averaged word-level gaze-duration dataset}, produced by averaging gaze durations across readers for each word (2,097 data points). Gaze duration is defined in \Cref{sec:previous-work-aggregations}.

\paragraph{Optical character recognition (OCR).}
We applied the Python Tesseract OCR library\footnote{\url{https://pypi.org/project/pytesseract/}} to each image in the MECO dataset \citep{siegelman2022expanding} to identify textual characters and their bounding boxes. 
The number of characters per text, the number of lines, and bounding box information are summarized in \Cref{tab:summary_bbox}.
Tesseract provides the position and dimensions of each recognized character. We set the heights to a constant value by adjusting them to match the tallest character in the image, and the widths to the 90$^{\text{th}}$ percentile of character widths. This ensures a consistent character grid for subsequent analysis.
Because Tesseract does not detect whitespace as distinct regions, we identified whitespace ourselves by comparing gaps between adjacent character boxes. Whenever the horizontal gap was at least 80\% of a typical single-character width, we considered the gap to be a whitespace and assigned it a bounding box of the same constant height.

\begin{table}[t]
\centering
\begin{tabular}{lrrrr}
\hline
 & Avg & SD & Min & Max \\
\hline
Lines & 10.5 & 1.2 & 8.0 & 12.0 \\
Characters & 1093.0 & 125.2 & 831.0 & 1231.0 \\
BBox width & 12.0 & 0.0 & 12.0 & 12.0 \\
BBox height & 22.3 & 2.2 & 18.0 & 24.0 \\
\hline
\end{tabular}
\caption{Summary statistics of raw MECO data, including the number of lines and characters per text and bounding box (BBox) dimensions.}
\label{tab:summary_bbox}
\end{table}

\paragraph{Model training and evaluation.} 
Our models were implemented in PyTorch and trained with gradient-based optimization using stochastic gradient descent \citep{robbins1951stochastic}, with Nesterov momentum \citep{nesterov1983method}. The models were trained on a fixed split of 80\% training, 10\% validation, and 10\% test data over 30 epochs, using early stopping with a patience of 5 epochs (i.e., training stopped if the validation loss did not improve for 5 consecutive epochs). The best hyperparameter configuration was selected based on validation performance and then evaluated on the held-out test set. For the models related to fixation onsets and locations, we performed a grid search over batch sizes $\{64, 128, 256\}$, learning rates $\{0.1, 0.01, 0.001\}$, and weight‑decay coefficients $\{0, 10^{-4}\}$, for a total of $18$ runs for each model and dataset type. For the convolutional model of fixation durations, the batch size was fixed at 128, while learning rates $\{0.01, 0.001, 0.0001\}$ and weight‑decay values $\{0, 10^{-4} \}$ were explored. We further varied the initial parameters of the Gamma distribution that defines the convolution kernel in \cref{eq:log_mean_duration}, testing $(\agamma,\bgamma)\in \{(2,3), (3,4), (3,6) \}$ and an initial $\thgamma$ of 0.5, for all predictors $k = \{ 1, \ldots, K \}$. These values were selected empirically so that the resulting distributions would have means that align with different inter‑arrival times (from $0.5$ to $0.75$ seconds). Durations were recorded in milliseconds, but onset times were rescaled to seconds before being passed to the convolution and saccade models. This scaling keeps inter‑arrival values within a range that avoids gradient explosion. We employed warm-starting to help in the search for a good set of parameter values. More specifically, we trained the models sequentially, starting with the simpler models used as baselines and using those for initialization in more complex models. For the more complex models, we initialized shared parameters with the best values from the corresponding simpler model. We also performed additional analyses using a linear model. For those analyses, we employed five-fold cross-validation: on each fold, the model was fit using 80\% of the data and evaluated on the remaining 20\%. Goodness of fit was quantified by the log-likelihood ratio (i.e., delta log-likelihood) with respect to predefined baselines, specified in \cref{sec:saccade-experiments} and \cref{sec:duration-model-experiments}. For experiments evaluated on the fixed test split, predictive uncertainty was estimated via bootstrap resampling of the test-set predictions.

\paragraph{Predictor values.}
Each fixation that lands in a character's bounding box can be adorned with linguistic attributes the character itself and the unit, e.g., the word, it belongs to.
We consider the following attributes:
(i) the character-level surprisal given the context, (ii) the word-level surprisal of the corresponding word, (iii) the number of characters in the corresponding word, i.e., its length, and (iv) the unigram surprisal, i.e., the log frequency, of the corresponding word.
For bounding boxes associated with whitespaces we can only compute character-level surprisal; fixations that do not land in any bounding box remain adorned with these attributes.\footnote{An alternative approach would be to assign each fixation to the predictor values associated with either the previous or the closest bounding box; we do not explore these options.\looseness=-1}
We distinguish character- and word-level surprisal as both have been considered as predictors in past work \citep{giulianelli-etal-2024-proper}.
To obtain the surprisal values, we must estimate $\overrightarrow{p}(\cdot \mid \ctx)$ from \cref{eq:surprisal} using an autoregressive language model. 
Because most modern language models learn a distribution over \emph{token} sequences, we use \citeposs{vieira-2024-token-to-char-lm} algorithm to convert the language models to the character level.
Word-level surprisal is obtained by summing the surprisal values of the subword tokens that comprise the word; this algorithm is correct when no subword token crosses a word boundary, which is the case in the language models we experiment with. 
Word-level surprisal estimates are derived from mGPT \citep{mgpt2024}, while character-level surprisal estimates are derived from GPT-2 \citep{radford2019language}.
Finally, word lengths and word frequencies are obtained from \citet{robyn_speer_2022_7199437}.

\section{Discussion on Related Work}
The below subsections give more context on aggregated eye-tracking measurements (\cref{appendix:aggregations-review}) and cognitive models of reading (\cref{sec:related-models}).

\subsection{Common Aggregations of Reading Data and their Interpretation}
\label{appendix:aggregations-review}

We defined first fixation duration, gaze duration, and total fixation duration in \cref{sec:previous-work-aggregations}.
These are generally thought to reflect progressively later stages of language processing \citep{inhoff1984stages,berzak-2023-eye}.
We describe them here, along with their standard interpretations.
First-fixation time, the duration of only the first fixation that lands on a word, is associated with word identification and lexical processing \citep{clifton-2007-eye,berzak-2023-eye} and tends to exhibit smaller surprisal effects \citep{wilcox-2023-testing,deVarda2024}.
Gaze duration (also called first-pass time), the summed duration of all fixations between landing on a word's region and leaving it, is thought to be indicative of early syntactic and semantic processing, and typically considered the aggregate to be most strongly associated with processing difficulty \citep[e.g.,][]{smith2013-log-reading-time,goodkind-bicknell-2018-predictive,wilcox2020}.
Total fixation time, the summed duration of all the fixations on the word, including refixations of the region after it was left, is thought to be indicative of integrative processes \citep[e.g.,][]{demberg-keller-2008-data,Roberts-Siyanova-Chanturia-2013} and sometimes exhibits, somewhat unexpectedly, stronger surprisal effects than first-fixation and first-pass time \citep{wilcox-2023-testing,giulianelli-etal-2024-proper}.
Scanpath durations, the summed duration of all \emph{consecutive} fixations on a word, are different from the above aggregated measurements in that they may contain several measurements for a given word if the word was fixated on multiple times, excluding consecutive re-fixations. Thus, fixation events are ordered according to the temporal order in which the words were fixated on, rather than the sequential order in which they are written out in the text. They have been modeled, e.g., by \citet{shain-schuler-2020-continuous}.

\subsection{Cognitive Models of Reading}\label{sec:related-models}

Several cognitive models of eye movement control during reading have been proposed. Among the most prominent are the E-Z Reader \citep{reichle2003ez} and SWIFT \citep{Engbert2005-vo}, which are probabilistic models of eye movements that can also be framed as point processes.
However, these models are not designed to capture the influence of linguistic processing on gaze behavior in a data-driven way. The Bayesian Reader \citep{Norris2006-NORTBR-3}, for instance, focuses on explaining effects such as the increased difficulty of processing low-frequency words.
Beyond these, other approaches have used machine learning to predict which words are fixated during reading, i.e., skip rates \citep{nilsson-nivre-2009-learning,hahn-keller-2016-modeling,wang2019new,bolliger-etal-2023-scandl}. Some models also predict fixation durations directly \citep{bicknell-levy-2010-rational,bolliger2025scandl2}.
In this article, we model the scanpaths at a more granular level, considering the exact spatial locations of fixations rather than just the words they land on. Although our model can learn effects that are related to both cognitive and oculomotor control processes, it is not meant to provide a plausible explanation of the mechanisms underlying such effects.

\section{Intensity Functions for Baseline Models}\label{app:baseline-models}

In this section, we provide more details on the Poisson and last-fixation baselines. Similarly to how we characterized the Hawkes process through \cref{eq:intensity-hawks}, we characterize each of these models through their intensity function. The density function from \cref{eq:hawkesdensity-ours} is changed accordingly.

\paragraph{Poisson process.}
The Poisson process model assumes that every fixation is equally likely regardless of spatial location or temporal history, which implies fixations that are independent in space and time. Its intensity function is
\begin{equation}
\intensity(\ntime, \location_n; \nphist) \defeq \baseintensity,
\end{equation}
where \(\baseintensity \in \R_{\geq 0}\) is the only learnable parameter. This serves as the simplest baseline, ignoring any spatial or temporal dependencies.

\paragraph{Last-fixation baseline.}
The last-fixation model introduces a basic spatial dependency by assuming that each new fixation is normally distributed around the most recent fixation. Let \(\location_{n-1}\) denote the location of the last fixation. Then, the intensity function is given by
\begin{equation}
\intensity(\ntime, \nloc; \nphist) \defeq \baseintensity + \spatdens_{n-1}\bigl(\nloc\bigr),
\end{equation}
where $\spatdens_{n-1}(\nloc)$ is the probability density function of a normal distribution centered at $\location_{n-1}$ with variance $\sigma^2$. This corresponds to the spatial distribution in \cref{eq:spatial_distr} when using $\spatmeanbase(\location) \defeq \location$, i.e., the identity function.\looseness=-1
 
\label{sec:appendix}
\section{Modeling Duration: Distribution Selection}
\label{appendix:lognormal}
In this section, we elaborate on our choice of the conditional density function for modeling fixation durations, as introduced in \cref{sec:duration-model}. We evaluated six candidate distributions commonly used for modeling right-skewed time-to-event data: Rayleigh, exponential, Weibull, normal, log-normal, and gamma. The evaluation was performed using 10-fold cross-validation on the training and validation data.
\begin{figure}
    \centering
    \includegraphics[width=0.6\linewidth]{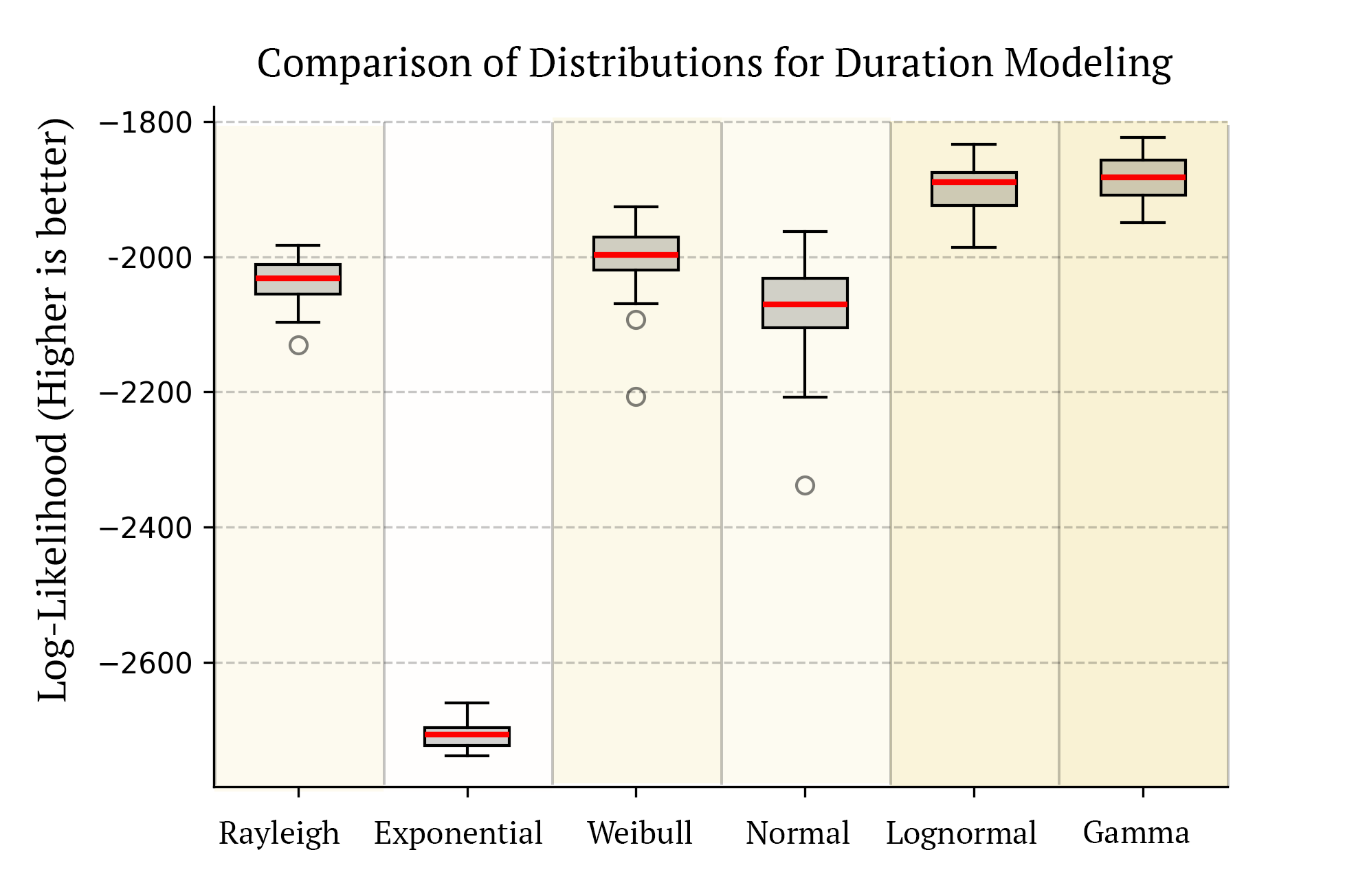}
    \caption{Goodness-of-fit comparison of candidate distributions for fixation durations. Both log-normal and gamma distributions showed superior performance compared to other alternatives; higher log-likelihood indicates better fit. The log-normal was ultimately selected for its enhanced parameter interpretability.}
    \label{fig:time-to-event-distributions}
\end{figure}
While the log-normal and gamma distributions demonstrated comparable predictive performance (see \cref{fig:time-to-event-distributions}), we selected the log-normal distribution as its parameters directly correspond to moments of the log-transformed durations, enabling more intuitive interpretation. Moreover, these parameters can also be interpreted as those of a normal distribution fitted to the log-durations, facilitating analysis through least squares estimators.

\section{Further Results}\label{sec:more-results}
\begin{figure*}[t]
    \centering
    \includegraphics[width=\textwidth]{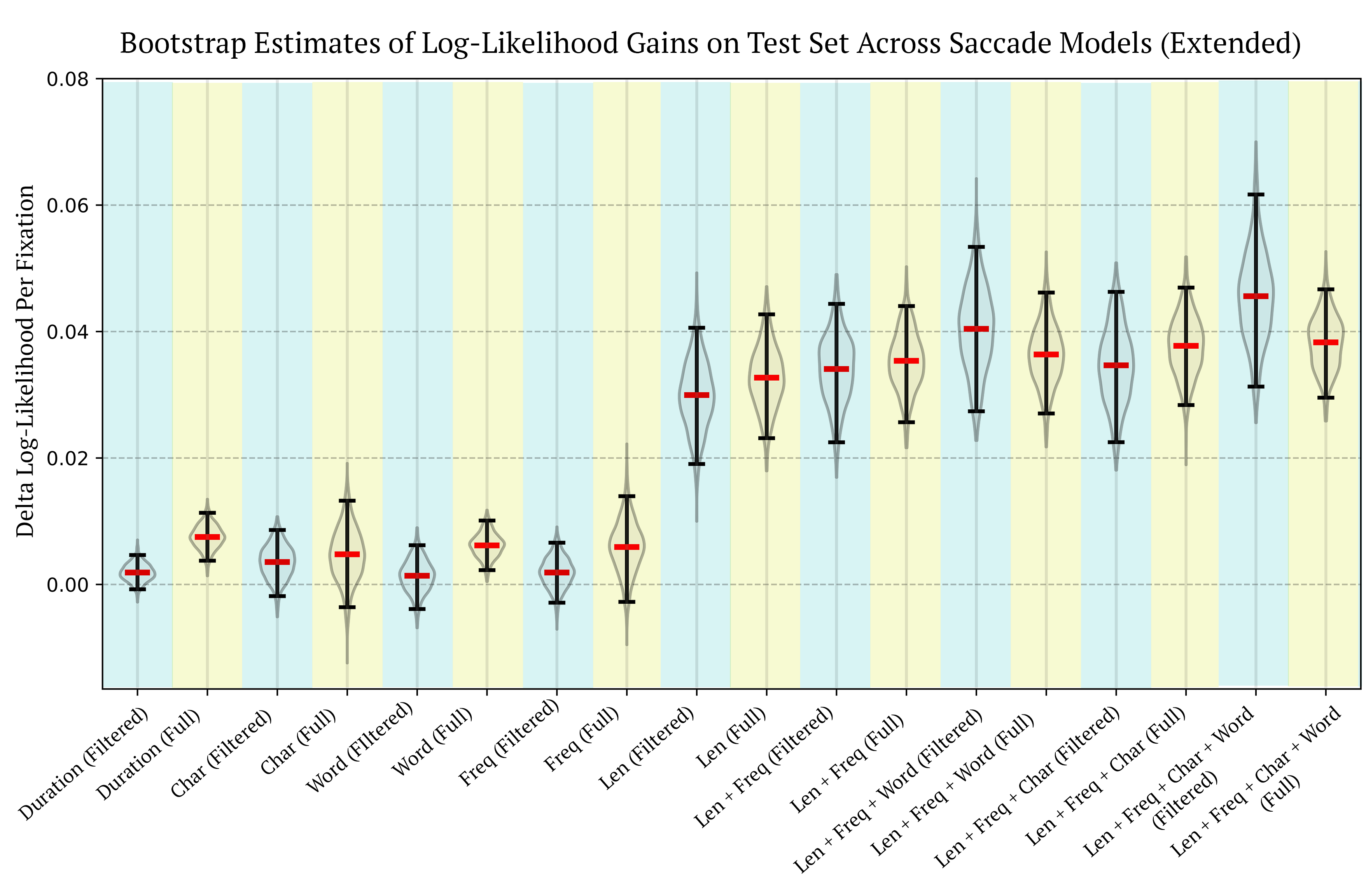}
    \vspace{-24pt}
    \caption{
Bootstrapped distributions of per-fixation log-likelihood gains across saccade models computed relative to the RSE model. The models correspond to those specified in \cref{sec:saccade-experiments}. The \emph{“Raw”} and \emph{“Filtered”} labels indicate whether the model was trained and evaluated on the raw or filtered scanpath dataset, respectively. We write ``word'' for word-level surprisal, ``char'' for character-level surprisal, and ``freq'' for unigram surprisal. Higher values indicate better predictive performance in modeling saccade durations. A shortened version of the plot can be found in \cref{fig:hawkes_process_results}.
    }
    \label{fig:hawkes_process_results_extended}
    \vspace{-2ex}
\end{figure*}

\begin{figure*}
    \centering
    \includegraphics[width=\textwidth]{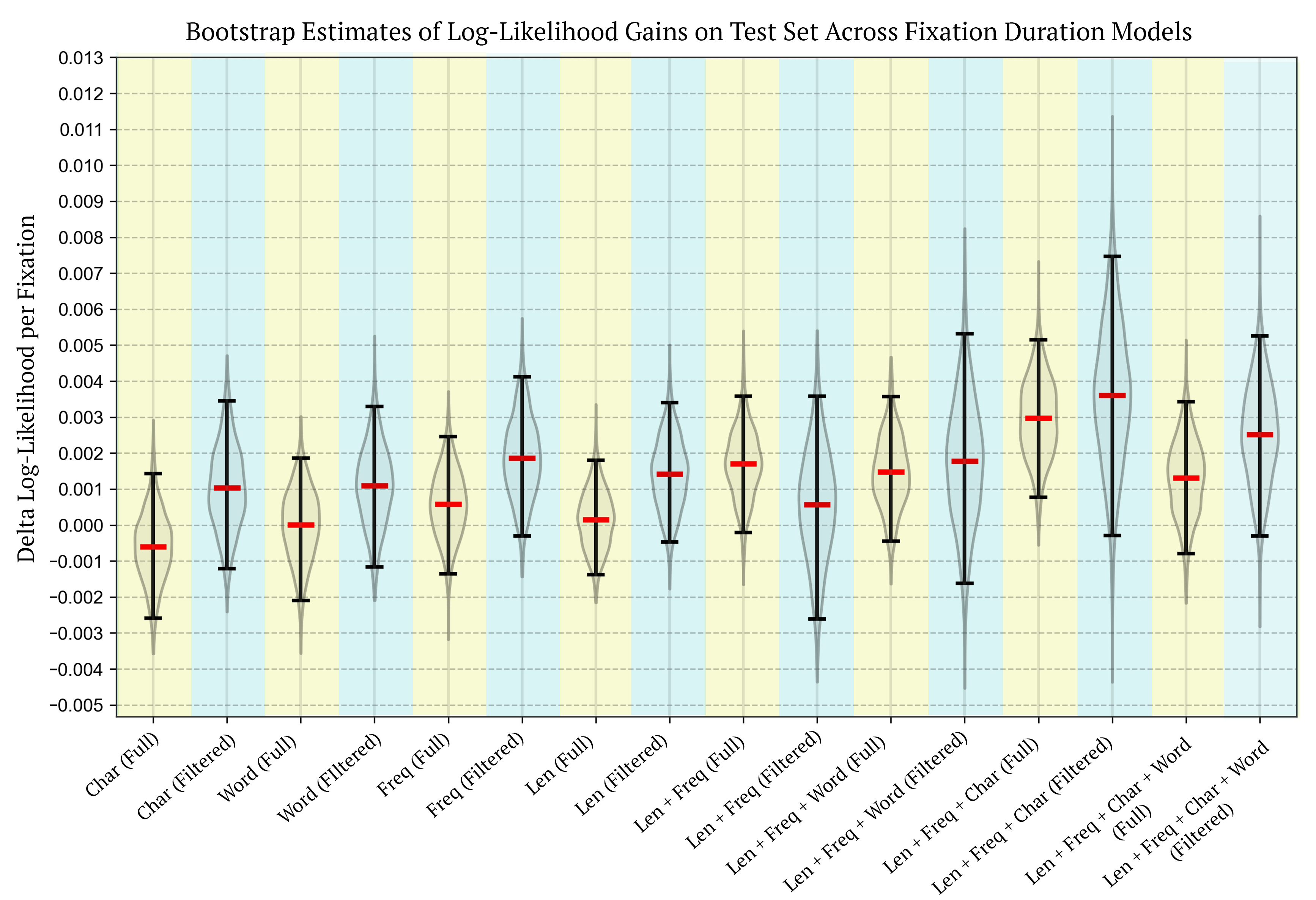}
    \caption{
Bootstrapped distributions of per-fixation log-likelihood gains across duration models, as defined in \cref{eq:log_mean_duration}. The base model includes reader-specific intercepts and duration spillovers via a convolution term. The full models incorporate the set of predictors indicated on the $x$-axis. For each predictor, the model includes its value at the current fixation, its interaction with the reader, and its spillovers through the convolution term. The models correspond to those described in \cref{sec:saccade-experiments}. The labels \emph{“Raw”} and \emph{“Filtered”} indicate whether the model was trained and evaluated on the raw or filtered scanpath dataset, respectively. We write ``word'' for word-level surprisal, ``char'' for character-level surprisal, and ``freq'' for unigram surprisal. For models trained on the raw dataset, an additional binary predictor indicates whether the fixation occurs on a word. Higher values reflect better predictive performance in modeling fixation durations.
}
    \label{fig:duration_convolution_results_figure}
    \vspace{-2ex}
\end{figure*}

\begin{figure*}[t]
    \centering
    \LARGE
    \includegraphics[width=\textwidth]{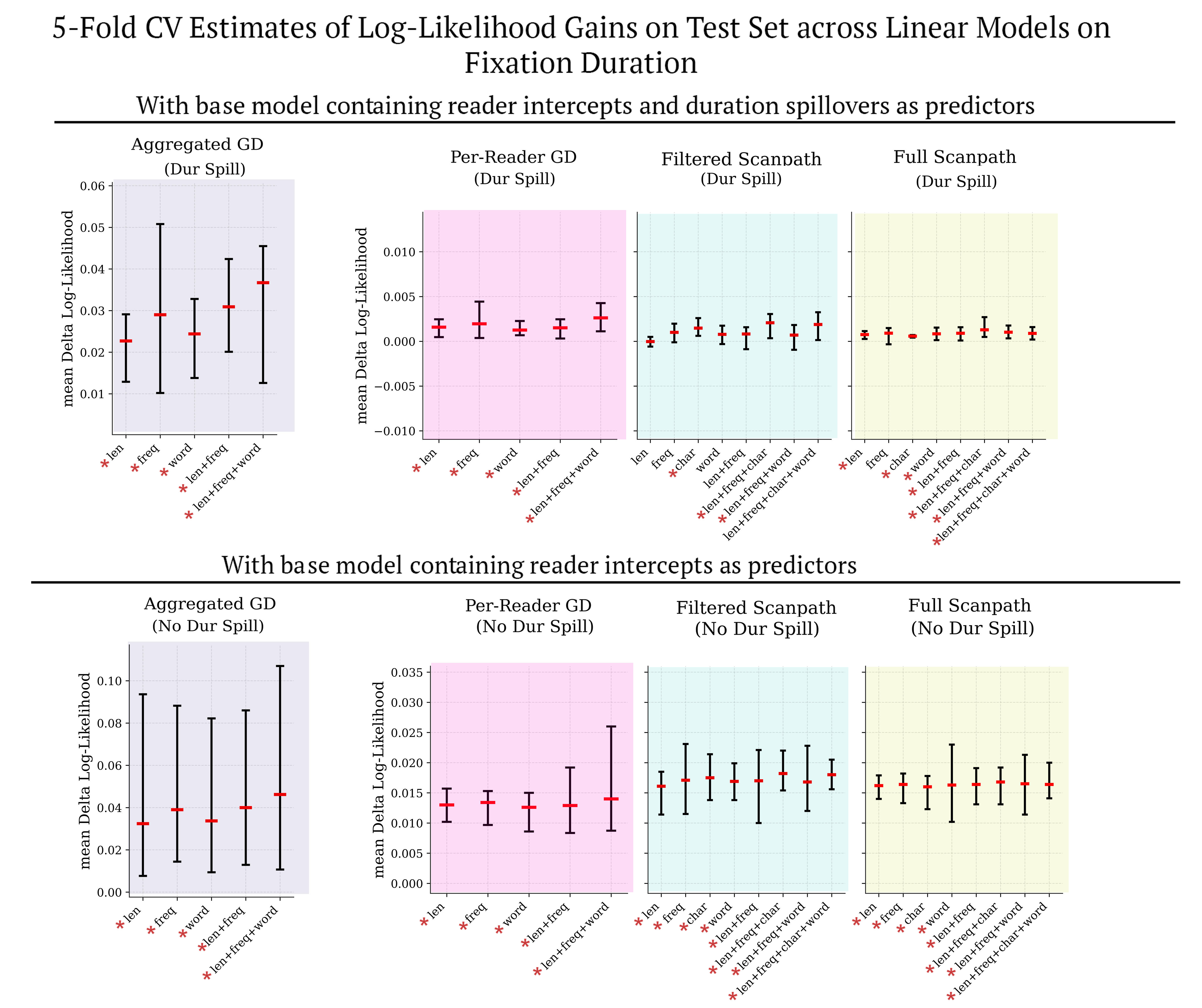}
    \vspace{-24pt}
    \caption{
Bootstrapped distributions of per-fixation log-likelihood gains across linear Markov models of fixation duration, as defined in \cref{eq:log_mean_duration_spillover_variables}.
Each delta log-likelihood is computed relative to a baseline model that includes: (i) reader-specific intercepts and spillover effects from previous fixation durations (models in the top panel of the figure), and (ii) only reader-specific intercepts (models in the bottom panel).
The models with spillovers include them from the previous two fixations. The $x$-axis labels specify the set of predictors included in each model variant. Each full model incorporates the specified predictors with corresponding spillovers, as well as the predictors in the corresponding baseline model. We write ``word'' for word-level surprisal, ``char'' for character-level surprisal, and ``freq'' for unigram surprisal. Gaze duration is abbreviated as ``GD''. 
Each plot corresponds to one of the datasets introduced in \cref{sec:experimental-setup}. 
Error bars represent the range—from minimum to maximum—observed across the five folds, with the central point indicating the arithmetic mean. We use asterisks (*) on the $x$-axis to highlight models that achieved a positive delta log-likelihood in all data folds.}
    \label{fig:linear_model_fix_duration_results}
    \vspace{-2ex}
\end{figure*}

In this section, we provide additional plots and analyses to complement the main results: \cref{fig:hawkes_process_results_extended} expands upon \cref{fig:hawkes_process_results}, \cref{fig:duration_convolution_results_figure} shows outcomes from the convolution model of fixation durations, and \cref{fig:linear_model_fix_duration_results} reports results for Markov linear models across various duration aggregation strategies.

\end{document}